\pdfoutput=1

\documentclass[11pt]{article}

\usepackage{emnlp2023}

\usepackage{graphicx}
\usepackage{booktabs}
\usepackage{array}
\usepackage{tabularx}
\usepackage{subcaption}
\usepackage{algorithm}
\usepackage{algpseudocode}
\usepackage{amsmath}

\usepackage{listings}
\usepackage{xcolor}

\usepackage{times}
\usepackage{latexsym}

\usepackage[T1]{fontenc}

\usepackage[utf8]{inputenc}
\usepackage{microtype}
\usepackage{inconsolata}

\title{Reflection-Based Memory For Web navigation Agents}

\author{
    Ruhana Azam \\
    UIUC \\
    \texttt{razam2@illinois.edu} \\\And
    Aditya Vempaty \\
    Emergence AI \\
    \texttt{aditya@emergence.ai} \\\And
    Ashish Jagmohan \\
    Emergence AI \\
    \texttt{ashish@emergence.ai}
  }

\begin{document}
\maketitle
\begin{abstract}
Web navigation agents have made significant progress, yet current systems operate with no memory of past experiences -- leading to repeated mistakes and an inability to learn from previous interactions. We introduce Reflection-Augment Planning (ReAP), a web navigation system to leverage both successful and failed past experiences using self-reflections. Our method improves baseline results by 11 points overall and 29 points on previously failed tasks. These findings demonstrate that reflections can transfer to different web navigation tasks.
\end{abstract}

\section{Introduction}
As language models have rapidly improved, so has the performance of agents that can tackle digital tasks such as web navigation \cite{zhou2023webarena, sodhiStePStackedLLM2024a, akkil2025agente}. Current web navigation systems are built by leveraging fixed data sets used for fine-tuning \cite{shen2024scribeagent, murty2024nnetnav, fuAutoGuideAutomatedGeneration2024} or in-context learning \cite{zhou2023webarena, murty2024nnetnav}. Although these methods have shown promising benefits, these agents still fail to iteratively improve over their life cycle. For example, when asked to \textit{find me a recipe on Allrecipes.com with 5 stars,} these systems repeatedly attempt to use a star filter that does not exist on the website. Recent works have attempted to resolve these issues by storing prior trajectories \cite{lutz2024wilbur, sarchVLMAgentsGenerate2025a}. In-Context Abstraction Learning (ICAL) addresses this challenge by iteratively improving in-context examples \cite{sarchVLMAgentsGenerate2025a}, while AgentWorkFlow parses prior trajectory sub-goals and stores them as an ongoing memory system \cite{wangAgentWorkflowMemory2024}. Other work, WILBUR~\cite{lutz2024wilbur}, stores summaries of any prior trajectories into the agent's memory.

We propose Reflection-Augmented Planning (ReAP), which stores reflections from prior experiences and retrieves relevant ones for new tasks. Unlike prior work, ReAP focuses on key insights rather the complete workflow, allowing us to effectively leverage successful and failed experiences.

In our paper, we test a web-agent's ability to leverage reflection on previously performed tasks and novel tasks. Our method demonstrates a 11-point performance increase on tasks the agent hasn't performed before, without additional training. Our findings demonstrate that simple key insights, rather than complete trajectory examples, sufficiently enhance agent performance and transfer effectively across diverse web tasks.



\section{Related Work}

\paragraph{Memory Augmented-Generation}
Augmenting language models with external memory has reduced hallucinations and improved generation reliability across various applications \cite{lewis2020retrieval, borgeaud2022improving, wang2024memoryllm}. Recent work has focused on not only storing documents or prior agent histories, but also condensed information in the form of high-level insights or the reasoning process.  
\cite{suzgunDynamicCheatsheetTestTime2025} demonstrates performance gains by maintaining a document of insights and strategies. Related approaches include Thought-Retriever \cite{feng2024thoughtretriever}, which logs models' chain-of-thought from past queries, and Buffer-of-Thoughts \cite{yangBufferThoughtsThoughtAugmented2024}, which distills high-level "thought templates" from problem-solving processes.
Among these, ReAP is conceptually most similar to \cite{suzgunDynamicCheatsheetTestTime2025}, though our work explicitly uses reflections towards web navigation. 

\paragraph{Self-Reflection} 
Self-reflection has significantly enhanced model performance across sequential decision-making tasks \cite{shinnReflexionLanguageAgents2023, madaanSelfRefineIterativeRefinement2023, gouCRITICLargeLanguage2024, anLearningMistakesMakes2024}, enabling systems to identify errors and refine their strategies. Recent studies have demonstrated its particular effectiveness for web navigation agents \cite{panAutonomousEvaluationRefinement2024, azamMultimodalAutoValidation2024, yang2024agentoccam}. Our work builds upon these findings by implementing a persistent memory system that captures and leverages reflective insights across multiple sessions and different tasks.

\section{ReAP Method}
The Reflection-Augmented Planning (ReAP) system operates as a specialized retrieval mechanism where the knowledge base contains reflections that help the agent perform more effectively.

\textbf{The memory} stores key-value pairs representing previously encountered trajectories. Keys contain user tasks (e.g., \textit{Search for a recipe with ``chicken breast'' and ``quinoa'' under 30 minutes on Allrecipes}), while values store reflections gained during navigation (e.g., \textit{Allrecipes.com does not have a filter for preparation time}).

\textbf{The ranking model} selects relevant reflections for a newly given task. It uses embeddings of the current task and entries in the knowledge base to rank the relevance of each reflection to the new task.


\subsection{The Pipeline} \label{sec:rap_full_pipeline}
Algorithm~\ref{alg:rap_pipeline} presents our two-phase ReAP pipeline. First, we build a memory index with trajectory reflections $R(\tau_j)$ embedded using text embeddings. During inference, we retrieve the $k=5$ most relevant reflections using cosine similarity and incorporate them into the agent's prompt. 

\begin{algorithm}
\caption{ReAP Pipeline}
\label{alg:rap_pipeline}
\begin{algorithmic}[1]
\Function{BuildMemory}{$\mathcal{D}_{train}$}
    \State $\mathcal{R} \leftarrow \emptyset$
    \For{$(t_j, \tau_j) \in \mathcal{D}_{train}$}
         \State $\mathcal{R} \leftarrow \mathcal{R} \cup \{(t_j, R(\tau_j), \text{Embed}(t_j))\}$
    \EndFor
    \State \Return $\mathcal{R}$
\EndFunction
\Function{ReAPPrompt}{$t_i, \mathcal{R}$}
    \State $q \leftarrow \text{Embed}(t_i)$ \Comment{Embed current task}
    \State $\{K_1,...,K_k\} \leftarrow \text{Retrieve}(q, \mathcal{R}, k)$ 
    \State \Return $\text{Agent}(t_i \oplus K_1 \oplus \cdots \oplus K_k)$
\EndFunction
\end{algorithmic}
\end{algorithm}

Details about our embedding model and similarity metrics can be found in Appendix~\ref{semantic_similarity_details}. Next we present the different implementations of $R$ testing in our evaluations.

\subsection{Knowledge Types in ReAP}
The reflection process in the ReAP system enables the agent to analyze its performance and learn from prior web navigation experience. We test three different methods that leverage both positive and negative prior experiences:

\begin{itemize}
    \item \textbf{One-shot \& Reward Label:} Directly inserts the trajectory and reward, e.g., $(s_1, a_1, s_2, a_2, \dots, r)$.
    
    \item \textbf{Summary \& Reward Label:} Prompts an LLM to produce a high-level summary of the plan, reward, and reason for failure.
    
    \item \textbf{Web-Reflection \& Reward Label:} These reflections follow the components below:
    \begin{enumerate}
        \item \textbf{Positive Reinforcement:} Which steps helped accomplish the plan?
        \item \textbf{Limited Functionality:} Were there any limitations in the site's capabilities?
        \item \textbf{Shortcuts:} Can the task be completed more efficiently?
        \item \textbf{Backtracking/Challenges:} Where did the prior plan fail or require rerouting?
        \item \textbf{Feedback:} Suggestions for improving an unsuccessful plan.
    \end{enumerate}
\end{itemize}

The exact prompts can be found in Appendix~\ref{app:prompts}.

\section{Hypothesis 1: ReAP can improve the agent's ability to redo prior tasks} \label{sec:h1}

In this section, we test whether the ReAP system can improve an agent’s performance on previously attempted tasks. WebArena is a simulated web environment containing shopping, mapping, Reddit, and GitLab interfaces that allows for standardized evaluation of web navigation agents. For this experiment, we use a set of 70 WebArena tasks~\cite{zhou2023webarena} and provide the agent with a single trajectory rollout for each task. The knowledge base contains trajectories from a vanilla run of our base agent, AgentOccam~\cite{yang2024agentoccam}, which uses \texttt{gpt-4o-2024-08-06}. We set the temperature to $0$ and top-$p$ to $0.5$ to increase determinism. Additional details on the model's configuration can be found in Appendix~\ref{app:temperature}.

\subsection{Success Rate (SR)}
To evaluate the effectiveness of the ReAP system in the context of similar tasks, we measured SR across multiple knowledge retrieval conditions. We compared baseline agent performance against the three aforementioned variants. The baseline results presented are from our rerun of AgentOccam.

\begin{figure}[ht]
    \centering
    \includegraphics[width=1\linewidth]{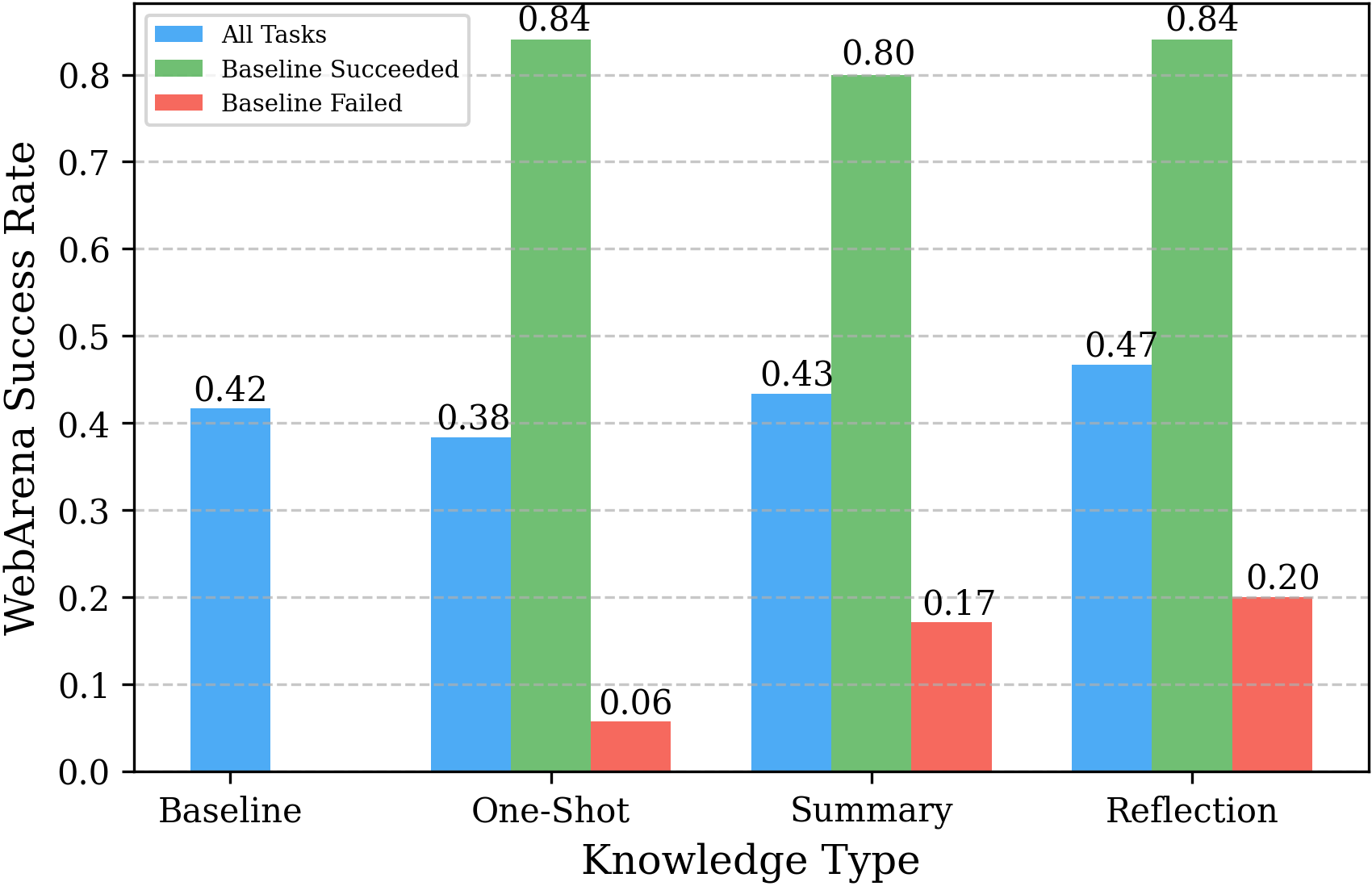}
    \caption{Success rate of 70 WebArena tasks. Results are split based on if the reflection was from a previously successful or unsuccessful trajectory.}
    \label{fig:h1_success_rates}
\end{figure}

The results in Figure~\ref{fig:h1_success_rates} reveal several important findings about reflection-based prompting:

\begin{itemize}    
    \item \textbf{Overall SR improvement:} All reflection methods increased the agent's performance over the baseline by +5 points.

    \item \textbf{Learning from failure:} Notably, agents improve by +20 points on previously failed tasks when provided reflections from previously failed trajectories. This demonstrates that reflections can prevent prior failure points.

    \item \textbf{Some Negative Transfer:} For previously successful tasks, none of the reflection methods achieved perfect success rates, with even the best methods (Summary and Reflection) only reaching 84\% SR. This indicates some negative transfer and suggests that ReAP works better for tasks that were previously difficult, although ReAP still maintains fairly high SR overall.
    

\end{itemize}

These results strongly support our hypothesis that ReAP systems can enhance an agent’s ability to redo tasks, regardless of whether the previous attempt was successful.

\subsection{Cost Analysis}

\begin{table}[!ht]
\centering
\small
\setlength{\tabcolsep}{3pt}
\begin{tabular}{@{}lrrrr@{}}
\toprule
\multicolumn{5}{c}{\textbf{Overall Performance (70 WebArena tasks)}} \\
\midrule
\textbf{Knowledge} & \textbf{Total Tok.} & \textbf{Prompt Tok.} & \textbf{Tct (s)} & \textbf{Steps} \\
\midrule
N/A & 221k & 202k & 682 & 11.92 \\
One-shot & 247k & 236k & 515 & 10.45 \\
Summary & 58k & 54k & 334 & 11.27 \\
Reflection & 225k & 205k & 556 & 10.08 \\
\midrule
\multicolumn{5}{c}{\textbf{Positive Examples Only}} \\
\midrule
N/A & 217k & 194k & 576 & 7.68 \\
One-shot & 243k & 225k & 572 & 6.60 \\
Summary & 42k & 39k & 275 & 9.12 \\
Reflection & 40k & 37k & 258 & 8.52 \\
\midrule
\multicolumn{5}{c}{\textbf{Negative Examples Only}} \\
\midrule
N/A & 224k & 208k & 757 & 14.94 \\
One-shot & 250k & 243k & 475 & 13.20 \\
Summary & 69k & 65k & 376 & 12.80 \\
Reflection & 358k & 324k & 770 & 11.20 \\
\bottomrule
\end{tabular}
\caption{Average cost for executing AgentOccam-Judge on 70 WebArena tasks using different knowledge types.}
\label{tab:h1_cost}
\end{table}

Beyond improving the success rate, ReAP significantly reduces the cost of running web navigation models. The number of steps is reduced by 15.4\% overall (from 11.92 to 10.08 with Reflection). Moreover, since failed tasks tend to produce long trajectories (average of 14.94 steps for the base agent), our method reduced the number of steps taken on these tasks by 25.0\% (to 11.20 steps).

We observed that the ReAP agent significantly reduced the need for re-prompting. The agent is re-prompted when invalid actions are generated. This can increase the token count. Due to this reduction in re-prompts and the overall decrease in steps, we observed substantial token and time reductions with certain approaches—particularly Summary, which cuts token usage by approximately 74\% (from 221k to 58k total tokens) and execution time by 51\% (from 682s to 334s). Overall, different ReAP approaches offer various tradeoffs when executing seen tasks. Summary provided the best token-count efficiency while Reflection offering the best SR while reducing the step count.
\section{Hypothesis 2: ReAP improves performance when similar tasks have been executed before} \label{sec:h2}

In this section, we show that the ReAP method extends beyond verbatim task repetition to facilitate transfer learning between similar tasks. To evaluate our hypothesis, we used the publicly available AgentOccam~\cite{yang2024agentoccam} trajectories for WebArena, dividing them into an 80-20 train-test split. We populated our knowledge base with reflections generated from the 80\% training trajectories and evaluated performance on the remaining 20\% of tasks. For each test task, we retrieved the top-$5$ reflections using cosine similarity and incorporated them into the agent's prompt. Appendix~\ref{semantic_similarity_details} provides further details on the semantic similarity measure used.

\subsection{Success Rates}
\begin{figure}[ht] 
    \centering
    \includegraphics[width=1\linewidth]{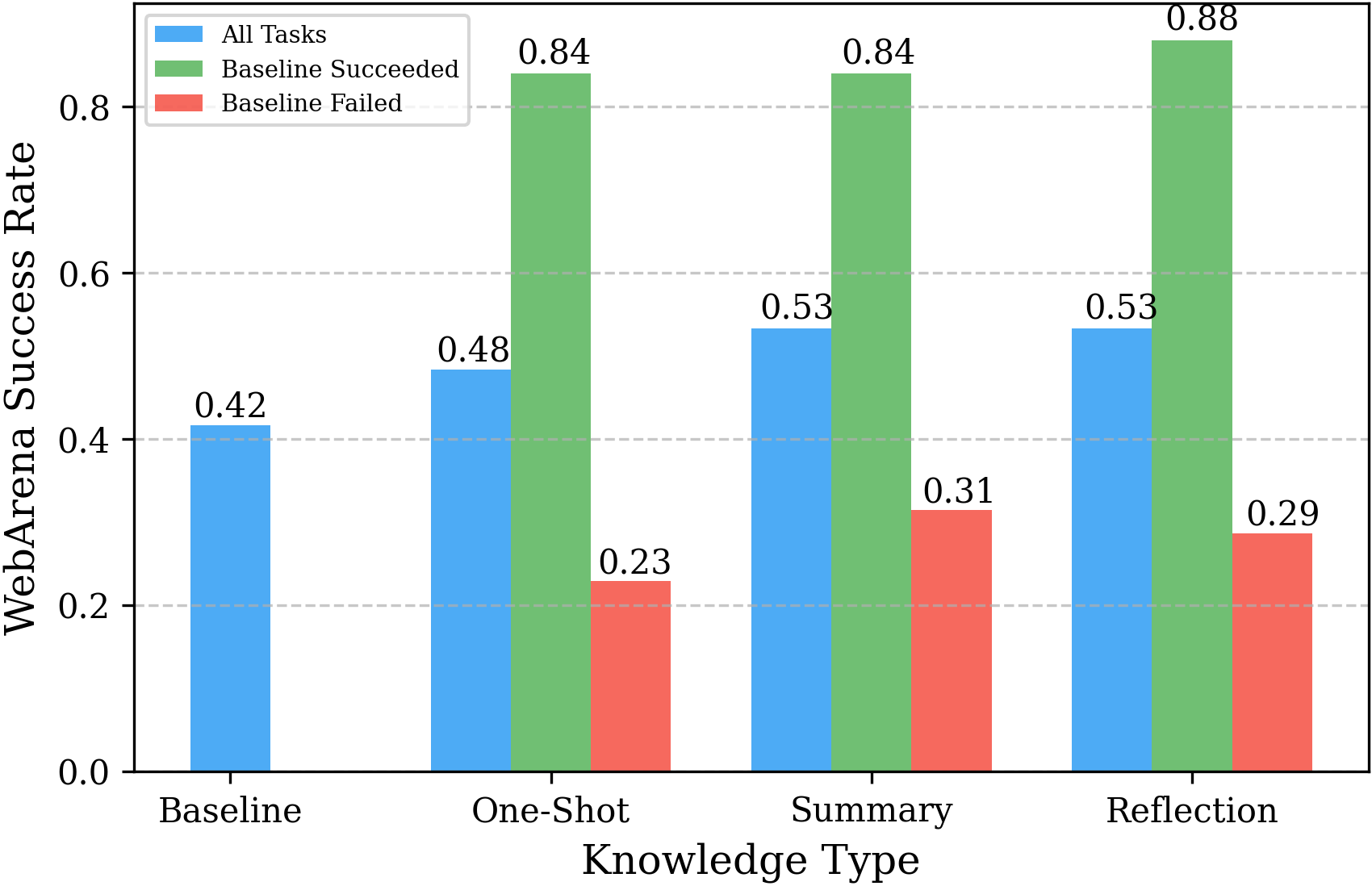}
    \caption{Success rate of 70 WebArena tasks. Results are split based on whether the baseline method succeeded or failed each task.}
    \label{fig:success_rates_h2}
\end{figure}

Figure~\ref{fig:success_rates_h2} illustrates performance improvements across different knowledge conditions. This plot shows the results for tasks that were previously failed or succeeded by the baseline agent—indicating which tasks had been historically difficult or easy.

The results in Figure~\ref{fig:success_rates_h2} reveal several important findings about knowledge transfer between similar tasks:

\begin{itemize}

\item \textbf{Overall Improvement in SR:} All knowledge types improved over the baseline, with One-shot achieving a 6 point improvement, while Summary and Reflection showing a 11 point improvement.

\item \textbf{Reflections benefit difficult tasks:} Most performance gains come from previously failed trajectories (28 points), demonstrating that reflections from similar tasks can effectively transfer and help solve challenging cases that the baseline agent could not complete.

\end{itemize}

Our agent shows an improvement of 11 points over the baseline, demonstrating that the ReAP system can effectively enhance task performance on similar—but distinct—tasks. These results show that ReAP enables effective knowledge transfer, improving the agent's general capabilities.

\subsection{Cost Analysis}
\begin{table}[!ht]
\centering
\small
\setlength{\tabcolsep}{3pt}
\begin{tabular}{@{}lrrrr@{}}
\toprule
\multicolumn{5}{c}{\textbf{Overall Performance (70 WebArena tasks)}} \\
\midrule
\textbf{Knowledge} & \textbf{Total Tok.} & \textbf{Prompt Tok.} & \textbf{Tct (s)} & \textbf{Steps} \\
\midrule
N/A (Our) & 221k & 202k & 682 & 11.92 \\
One-shot & 770k & 759k & 424 & 9.88 \\
Summary & 151k & 147k & 372 & 9.55 \\
Reflection & 83k & 80k & 327 & 8.45 \\
\midrule
\multicolumn{5}{c}{\textbf{Positive Examples Only}} \\
\midrule
N/A (Our) & 217k & 194k & 576 & 7.68 \\
One-shot & 314k & 311k & 271 & 7.60 \\
Summary & 71k & 68k & 239 & 7.48 \\
Reflection & 70k & 67k & 258 & 7.12 \\
\midrule
\multicolumn{5}{c}{\textbf{Negative Examples Only}} \\
\midrule
N/A (Our) & 224k & 208k & 757 & 14.94 \\
One-shot & 1095k & 1079k & 533 & 11.51 \\
Summary & 208k & 203k & 467 & 11.03 \\
Reflection & 92k & 89k & 377 & 9.40 \\
\bottomrule
\end{tabular}
\caption{Average cost for executing AgentOccam-Judge on 70 WebArena tasks using different knowledge types.}
\label{tab:h2_cost}
\end{table}

As in Section~\ref{sec:h1}, we find that ReAP reduces task execution cost while improving SR. The number of steps is reduced by 29.1\% overall compared to the baseline, with a particularly notable reduction of 37.1\% in steps taken on previously failed tasks.

Again, we observed that our approach significantly reduced the need for re-prompting, leading to a reduction in token usage. The reflection method in particular demonstrated the greatest efficiency gains, requiring only 83k total tokens compared to 221k for the baseline. Overall, our agent achieved substantial reductions in both steps and prompt tokens, with the reflection approach lowering steps by 29.1\%.

\section{Conclusion}
ReAP significantly improves web navigation tasks, increasing success rates by +11 points and decreasing the average number of steps by 34.7\%, thereby reducing the cost of running such agents. These results highlight a key finding: ReAP enables effective knowledge transfer between similar but distinct tasks, enhancing overall agent performance without requiring additional training. We hope that these advances in reflection-based memory will help make web agents more practical for everyday use by reducing errors and operational costs.


%

\section*{Ethics Statement}
Our web agent research could be misused to harvest sensitive data, bypass authentication systems, and automate fraud across websites. Malicious actors could exploit these tools to violate terms of service through mass scraping, credential stuffing, or denial-of-service attacks. Any deployment of automated web agents inherently risks digital ecosystem disruption through unintended traffic patterns or circumvention of anti-bot measures. We intend to further this research for beneficial applications and discourage implementation of our methods for malicious purposes that violate privacy, security, or digital system integrity.

\section*{Limitation}
The primary limitation of this study is our evaluation being bounded to the WebArena benchmark. While WebArena offers a controlled environment for benchmarking, extending evaluation to more real-life benchmarks such as Mind2Web~\cite{deng2023mind2webgeneralistagentweb} and WebVoyager~\cite{he2024webvoyagerbuildingendtoendweb} would provide a more realistic assessment of ReAP. However, WebVoyager requires human annotators and Mind2Web's automated evaluation system operates at approximately 85\% accuracy. Our resource limitations do not allow for human annotations, thus our benchmarks are limited. Another limitation is our experiments focused on a single base-agent. Examining ReAP's performance with open-source models would enhance accessibility for researchers with varied computational resources. Future research replicating our methodology across diverse models and benchmarks would strengthen our findings.


\bibliography{references}
\bibliographystyle{acl_natbib}

\appendix
\addcontentsline{toc}{section}{Appendix}
\clearpage
\onecolumn
\section{Prompt Templates by Knowledge Type} \label{app:prompts}
\subsection{Knowledge-Extraction Prompts}

\subsubsection{Summary Prompt}
\begin{lstlisting}
You are summarizing a prior web navigation episode from an autonomous agent.
Please answer the following based on the trajectory. Write out answers coherently as if you are passing along this information to an inexperience web navigation agent.

OBJECTIVE: {objective}

SUCCESSFUL?: {is_success}

SUMMARY GUIDELINES:
- What did the agent attempt to do?
- If successful, what were the key actions that worked?
- If unsuccessful, where did it go wrong and why?
- What parts of the environment were especially tricky?
- What should future agents be aware of on this type of website?

TRAJECTORY SNAPSHOT: {trajectory}
\end{lstlisting}

\subsubsection{Web-Reflection Prompt}
\begin{lstlisting}
You are analyzing a prior web navigation episode from an autonomous agent. Your goal is to output strategic advice and lessons learned to help an inexperienced agent perform better on similar websites or task in the future. Please answer the following based on the trajectory and objective shown:

1. Useful Subgoals: Which subgoals were necessary to accomplish the overall plan and sucessfully accomplished? For each sucessfully accomplished subgoal, what actions were taken to accomplish this?

2. Backtracking & Unexpected Challenges Faced: It's possible that there unexpected challenged while executing the objective. This can lead to unnecessary steps taken while executing the objective. If there was any backtracking, can you provide advice on how the backtracked steps can be avoided and how the task can be accomplished directly? Please do not provide any advice which you did not witness in the given trajectory.

3. Limited Functionalities Learned: Based on the trajectory, the agent may have learned there are certain capabilities not possible on certain websites (i.e. allrecipes.com has no option to filter recipes by rating). If any, what limitations does this current website have? 

4. Shortcuts Suggestions: Can you suggest any shortcuts for accomplishing the objective with fewer steps? Please only provide shortcuts you are certain are possible on the given website. If any shortcuts are suggested, please write them in a list.

5. Other Feedback: If the prior plan did not work, where did the agent go wrong? Please be specific. Do you have any tips which can help a inexperience agent avoid making the same mistake(s)?

Please write out answers is if you are speaking directly to the inexperience web navigation agent. Note the agent cannot see the full trajectory that you are viewing, so please answer accordingly. In each category, provide the answers as a bullet point list. Please do not call the agent an inexperienced -- that is rude.

OBJECTIVE: {objective}

SUCCESSFUL?: {is_success}

TRAJECTORY SNAPSHOT: {trajectory}
\end{lstlisting}

\twocolumn

\subsection{Agent's Objective Prompt}

\subsubsection{With Prior Example}
\begin{lstlisting}
Here are some examples of prior web interactions. Each example shows the task, the steps taken, and the outcome.

### Example {task_id} (Reward: {reward})
OBJECTIVE: {objective}
{trajectory}

### NEW TASK
OBJECTIVE:
{objective}
\end{lstlisting}

\subsubsection{With Summary}
\begin{lstlisting}
Here are some information from prior web interactions and their outcome.
Example {task_id}:
Task: {objective}
Outcome: {was_success}
Summary: {summary_str}

\end{lstlisting}

\subsubsection{With Reflection}
\begin{lstlisting}
Here are some tips from prior web interactions and their outcome.
Example {task_id}:
Task: {objective}
Outcome: {was_success}
Key Learnings: {reflect_str}
\end{lstlisting}

\section{Additional Results}
\subsection{Temperature Effect} \label{app:temperature}
Our experiments show that adjusting temperature and top-$p$ parameters impacts agent performance under the ReAP method. We evaluated performance across 37 WebArena tasks, comparing stochastic settings (temperature = 0.5, top-$p$ = 0.95) with more deterministic ones (temperature = 0, top-$p$ = 0.5).

\begin{table}[!ht]
\centering
\small
\begin{tabular}{@{}p{3.5cm}ccc@{}}
\hline
\textbf{Reflection Type} & \textbf{All} & \textbf{Pos.} & \textbf{Neg.} \\
\hline
Our Baseline & 0.32 & - & - \\
\hline
One-shot (Stochastic) & 0.28 & 0.67 & 0.11 \\
One-shot (Deterministic) & 0.47 & 0.92 & 0.28 \\
\hline
Summary (Stochastic) & 0.36 & 0.78 & 0.16 \\
Summary (Deterministic) & 0.47 & 1.00 & 0.16 \\
\hline
Reflect (Stochastic) & 0.36 & 0.78 & 0.16 \\
Reflect (Deterministic) & 0.51 & 1.00 & 0.28 \\
\hline
\end{tabular}
\caption{Success rates (SR) on 37 WebArena tasks with various temperature and top-$p$ settings.}
\label{tab:miniweb}
\end{table}

For both One-shot and Reflection approaches, more deterministic settings led to improved overall performance. The deterministic Reflection approach achieved the highest success rate (0.51) and perfect performance (1.00) on positive examples. This suggests that lower temperature values help agents better adhere to the trajectory information provided in prompts, improving their ability to follow demonstrated paths during task execution.
\subsection{Site-Specific Success Rates}
In this section, we examine site-specific success rates for the results presented in Sections~\ref{sec:h1}–\ref{sec:h2}.

\begin{figure}[ht!]
    \centering
    \includegraphics[width=0.3\textwidth]{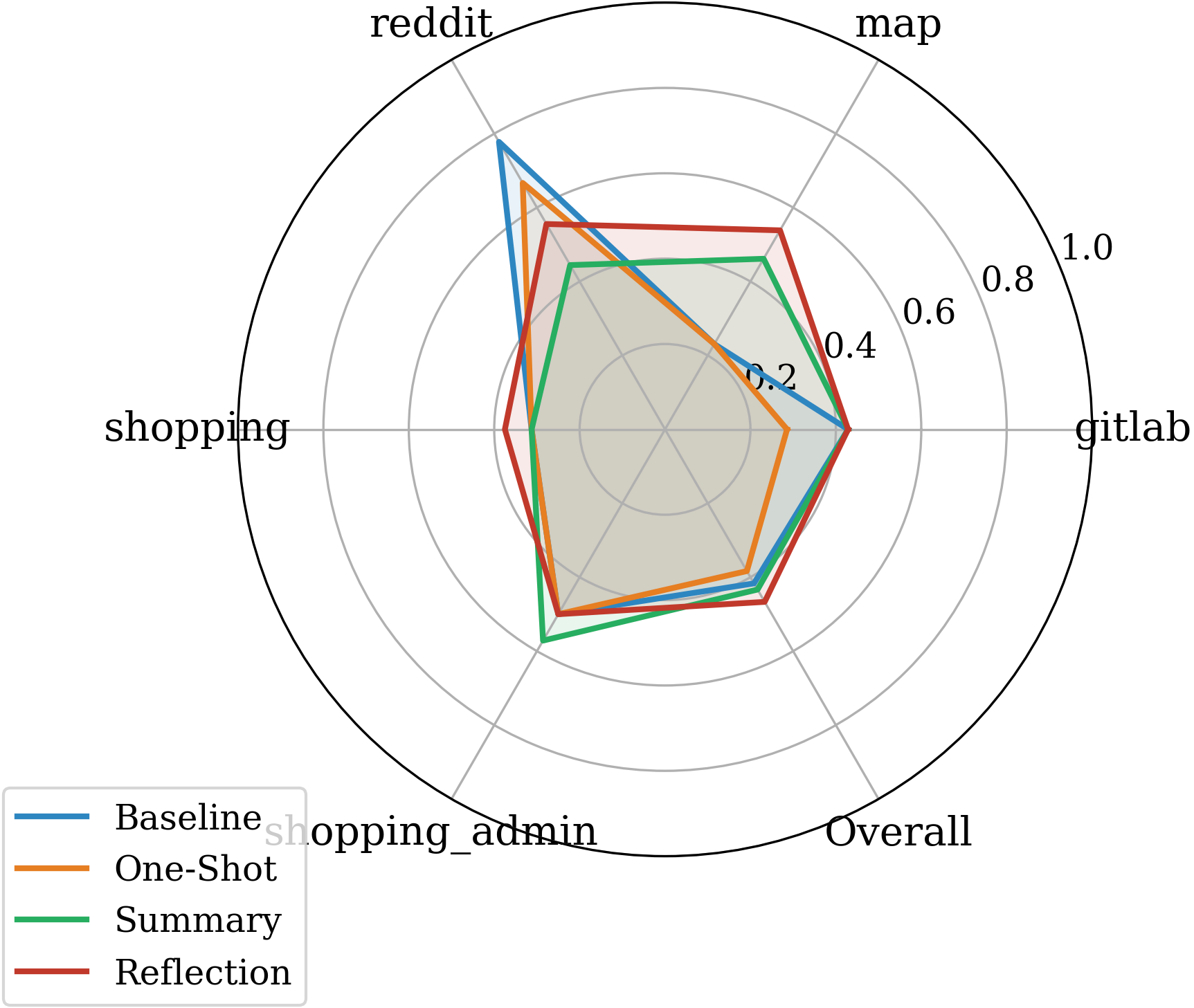}
    \caption{Success rate across 70 WebArena tasks, grouped by site, leveraging reflection for previously seen tasks (results from Section~\ref{sec:h1}).}
    \label{fig:site_spider_h1}
\end{figure}

For previously seen tasks (Figure \ref{fig:site_spider_h1}), we observe mixed results across websites. Map interfaces show substantial improvement with Reflection (30.7 points), while Reddit tasks demonstrate a notable performance decrease (22.2 points). Overall, Reflection achieves a modest 5.0 points improvement for previously seen tasks.

\begin{figure}[ht!]
    \centering
    \includegraphics[width=0.3\textwidth]{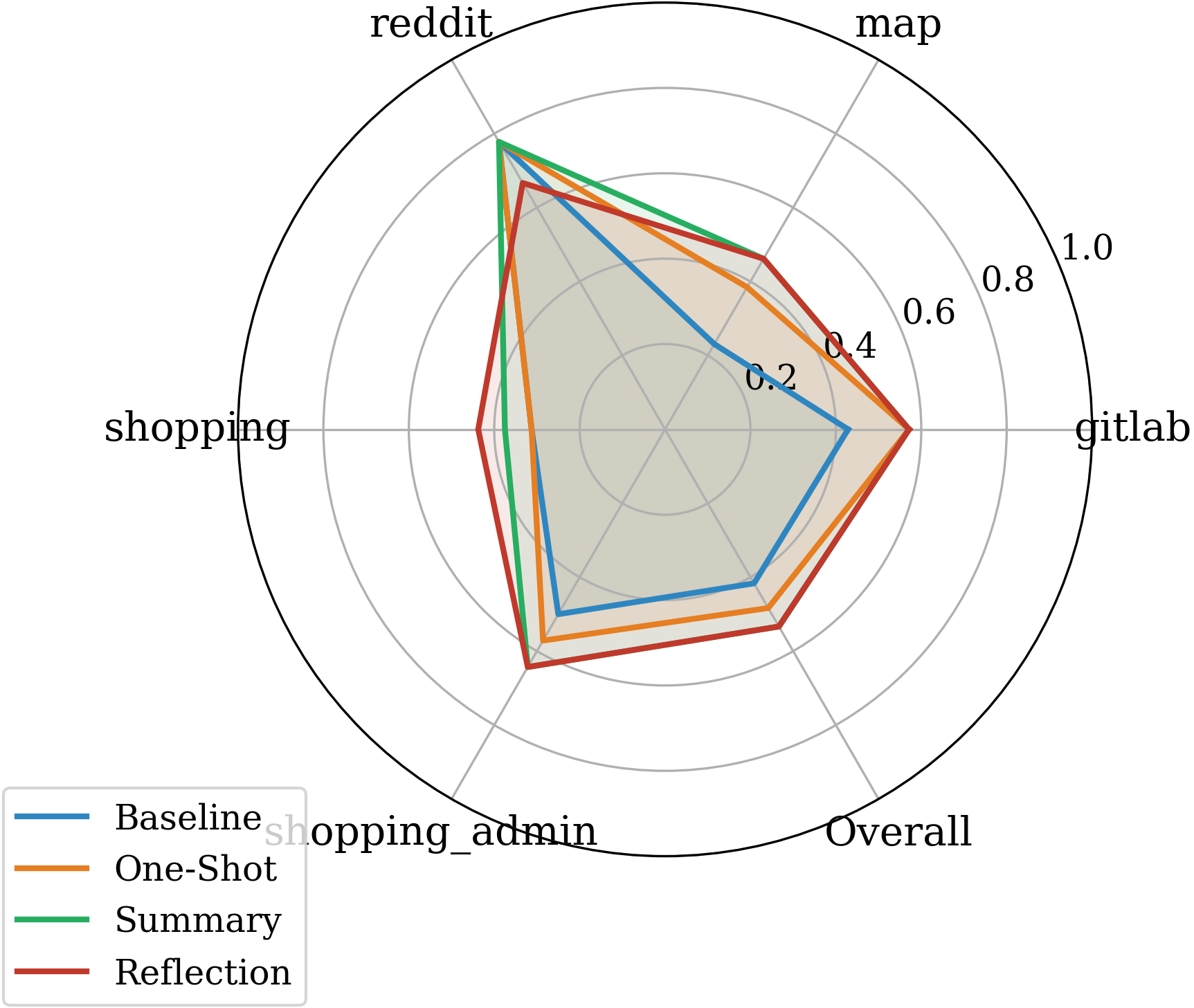}
    \caption{Success rate across 70 WebArena tasks, grouped by site, leveraging reflection for previously unseen tasks (results from Section~\ref{sec:h2}).}
    \label{fig:site_spider_h2}
\end{figure}

For unseen tasks (Figure \ref{fig:site_spider_h2}), improvements are more consistent across sites. Reflection outperforms the baseline by double-digit margins on most websites: 14.2 points for gitlab, 23.1 points for map, 12.6 points for shopping, and 14.3 points for shopping\_admin. Only reddit shows a performance decrease (11.1 points). These findings demonstrate that ReAP consistently benefits novel tasks across diverse web environments.
\begin{figure*}[ht!]
    \centering
    \begin{subfigure}[b]{0.3\textwidth}
        \centering
        \includegraphics[width=\linewidth]{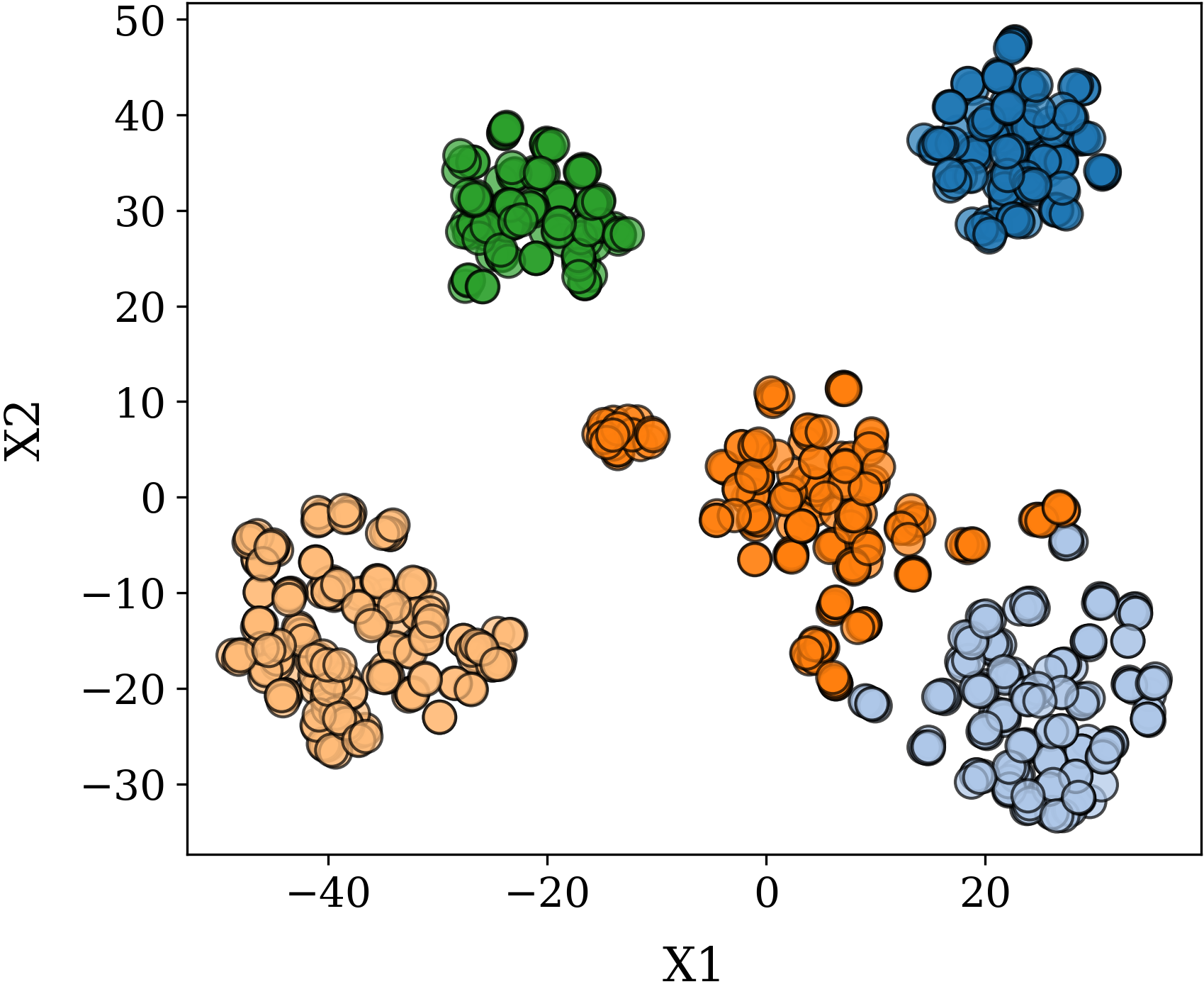}
        \caption{gte-Qwen2-7B-instruct}
        \label{fig:qwen-tsne}
    \end{subfigure}%
    \hspace{0.03\textwidth}%
    \begin{subfigure}[b]{0.3\textwidth}
        \centering
        \includegraphics[width=\linewidth]{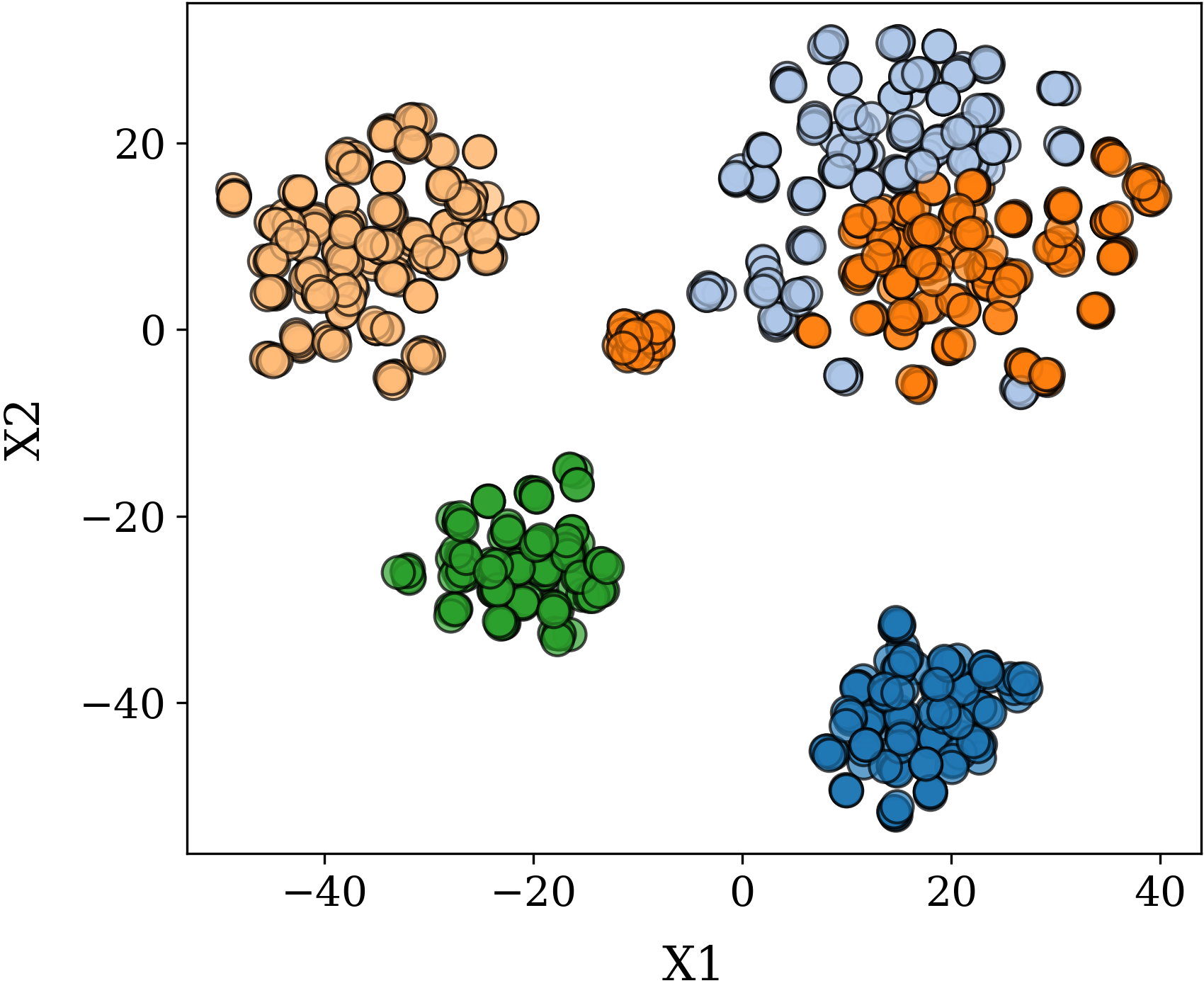}
        \caption{llama8b-nnetnav-wa}
        \label{fig:nnetnav-tsne}
    \end{subfigure}%
    \hspace{0.03\textwidth}%
    \begin{subfigure}[b]{0.3\textwidth}
        \centering
        \includegraphics[width=\linewidth]{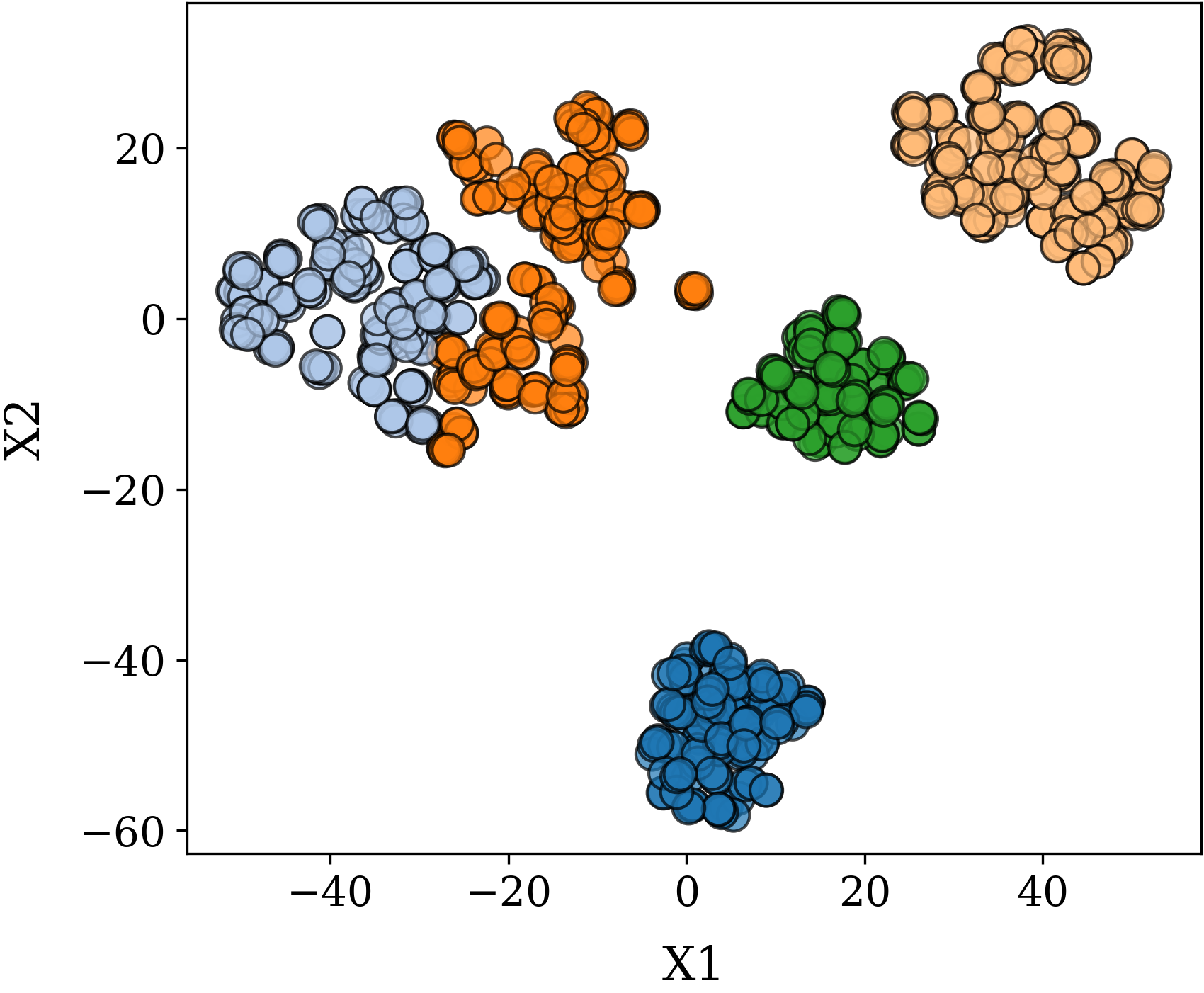}
        \caption{text-embedding-3-large}
        \label{fig:openai-large-tsne}
    \end{subfigure}
    
    \vspace{0.2cm}
    
    \begin{subfigure}[b]{0.3\textwidth}
        \centering
        \includegraphics[width=\linewidth]{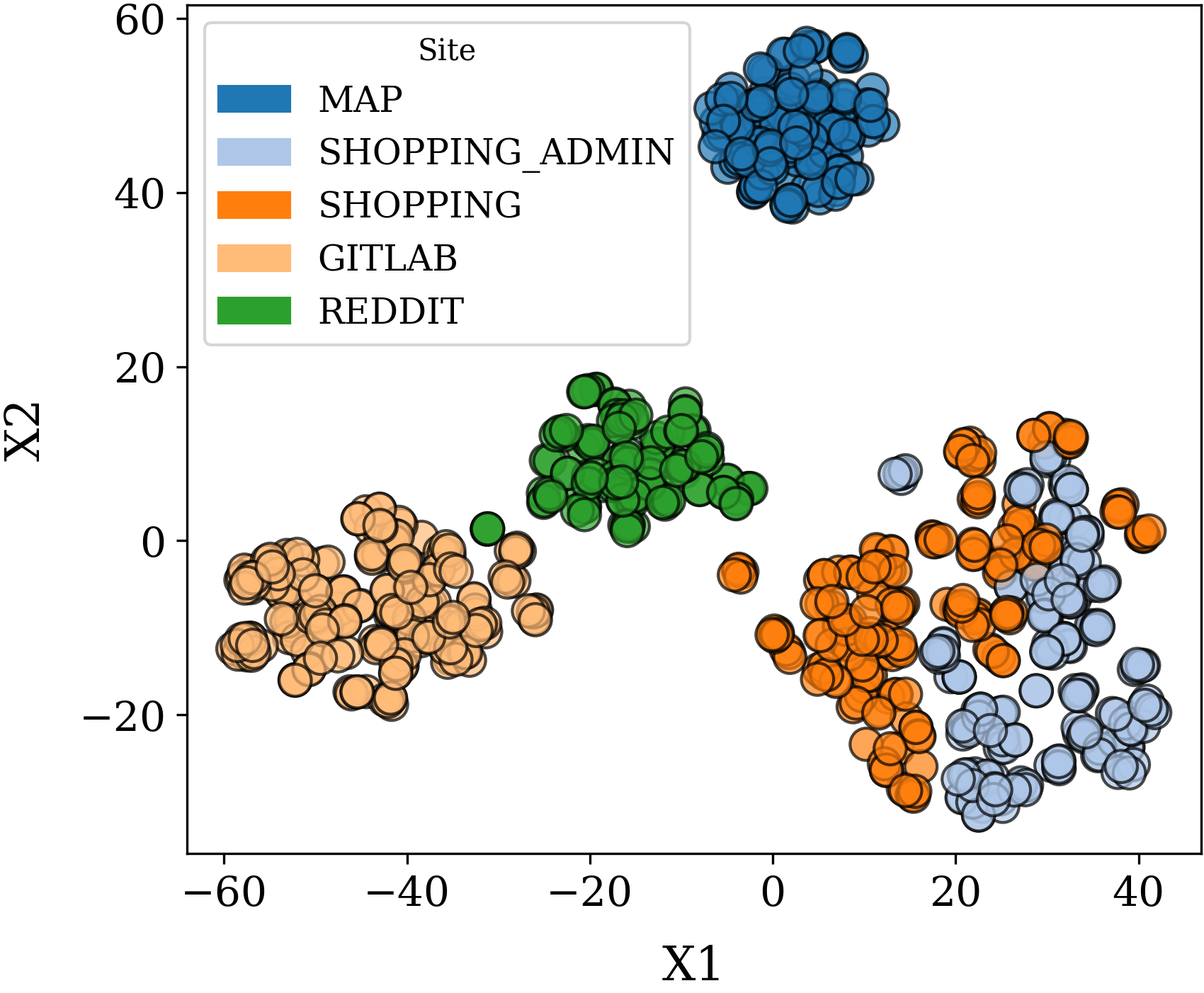}
        \caption{text-embedding-3-small}
        \label{fig:openai-small-tsne}
    \end{subfigure}%
    \hspace{0.03\textwidth}%
    \begin{subfigure}[b]{0.3\textwidth}
        \centering
        \includegraphics[width=\linewidth]{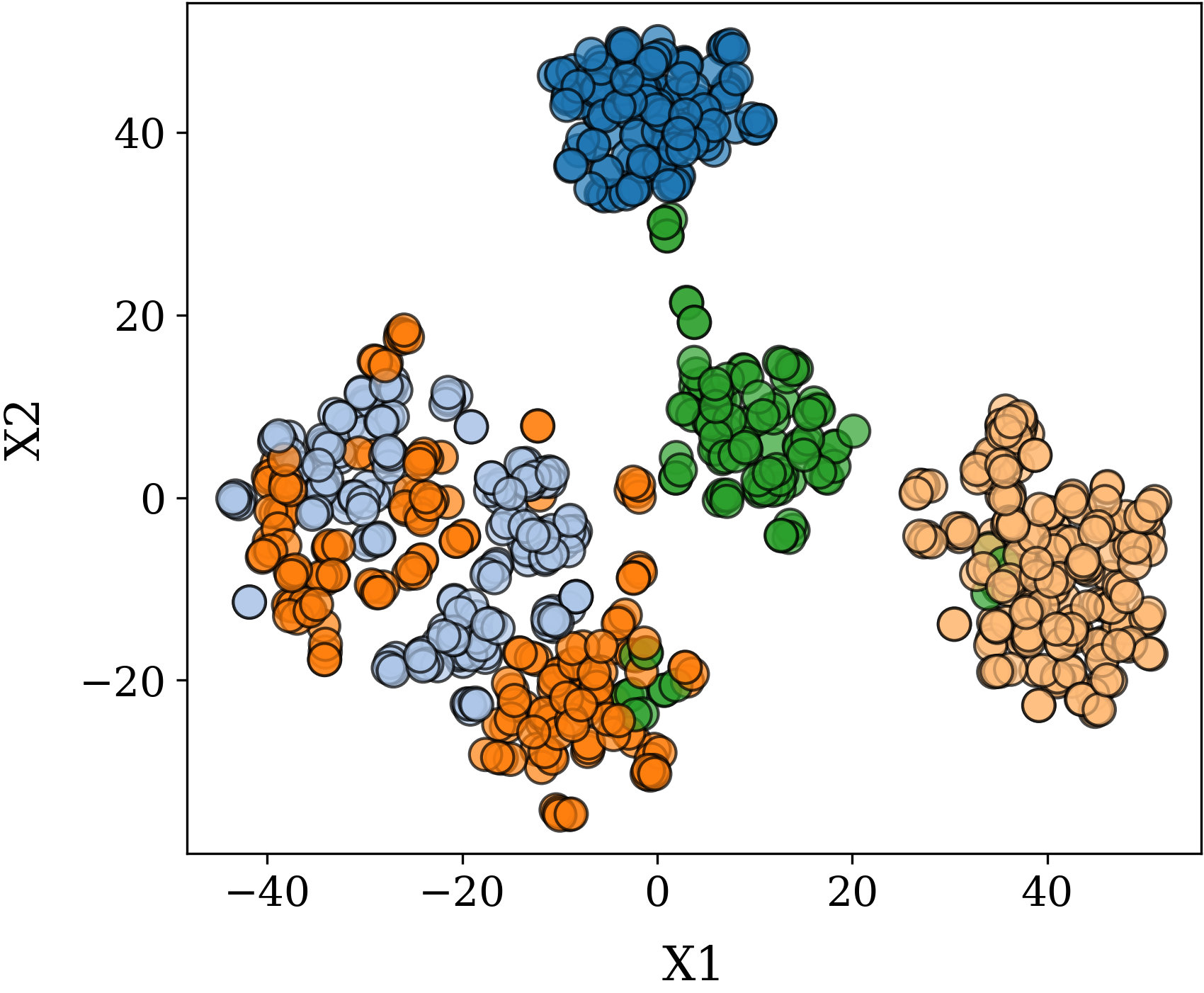}
        \caption{all-MiniLM-L6-v2}
        \label{fig:all-MiniLM-L6-v2-tsne}
    \end{subfigure}
    
    \caption{t-SNE visualization of task embeddings for WebArena tasks across different models.}
    \label{fig:tsne-embedding}
\end{figure*}

\section{Semantic Similarity for Task Retrieval} \label{semantic_similarity_details}

We retrieve relevant examples using semantic similarity between task instructions. Our method computes cosine similarity between task embeddings, where each task is represented as: \textit{"Go to \{start\_url\}, \{intent\}"} (e.g., \textit{Go to MAP, Which US states border Illinois?}). The primary goal is to assign high similarity scores to semantically related tasks.

We evaluate several embedding models: the domain-specific \texttt{llama8b-nnetnav-wa}~\cite{murty2024nnetnav}, the general-purpose \texttt{gte-Qwen2-7B-instruct}, OpenAI's \texttt{text-embedding-3} models (large/small), and the lightweight \texttt{all-MiniLM-L6-v2}. We were able to run local models on a Quadro RTX 8000 with 48GB of VRAM. A visualization of task embeddings over the full WebArena dataset is shown in Figure~\ref{fig:tsne-embedding}.

Ideally, tasks from the same website should share similar workflows and thus cluster together in the embedding space. Figure~\ref{fig:tsne-embedding} shows that smaller models such as \texttt{text-embedding-3-small} and \texttt{all-MiniLM-L6-v2} struggle to make this separation—particularly between \texttt{shopping} and \texttt{shopping\_admin} tasks.

\subsection{Semantic Similarity Measure}
We evaluated embeddings on a set of 30 WebArena tasks, with a focus on distinguishing closely related categories (e.g., \texttt{shopping} vs. \texttt{shopping\_admin}).

\begin{figure*}
    \centering
    \includegraphics[width=0.75\textwidth]{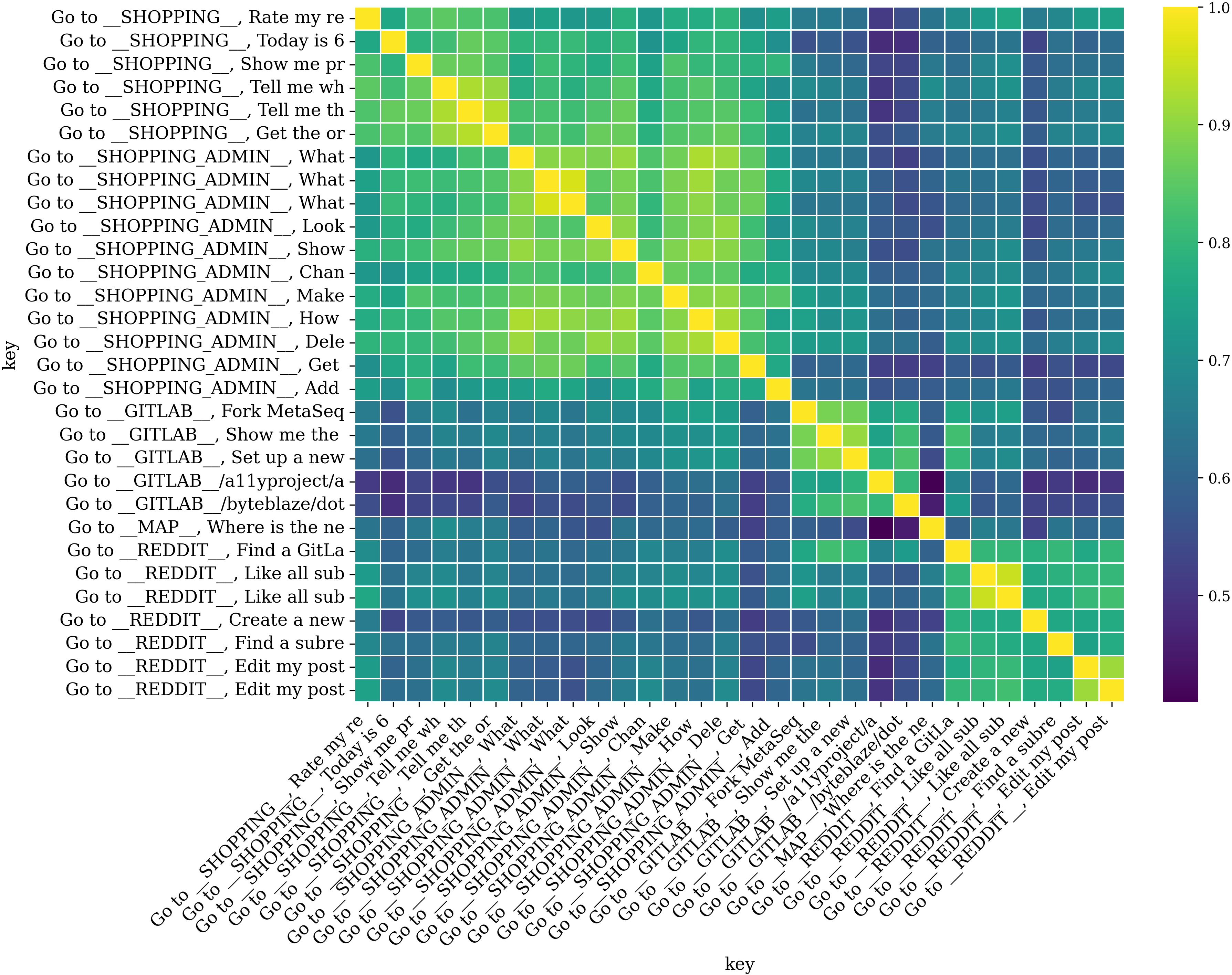}
    \caption{Cosine similarity heatmap for \texttt{gte-Qwen2-7B-instruct} embeddings across WebArena task categories.}
    \label{fig:qwen-heatmap}
\end{figure*}

\begin{figure*}
    \centering
    \includegraphics[width=0.75\textwidth]{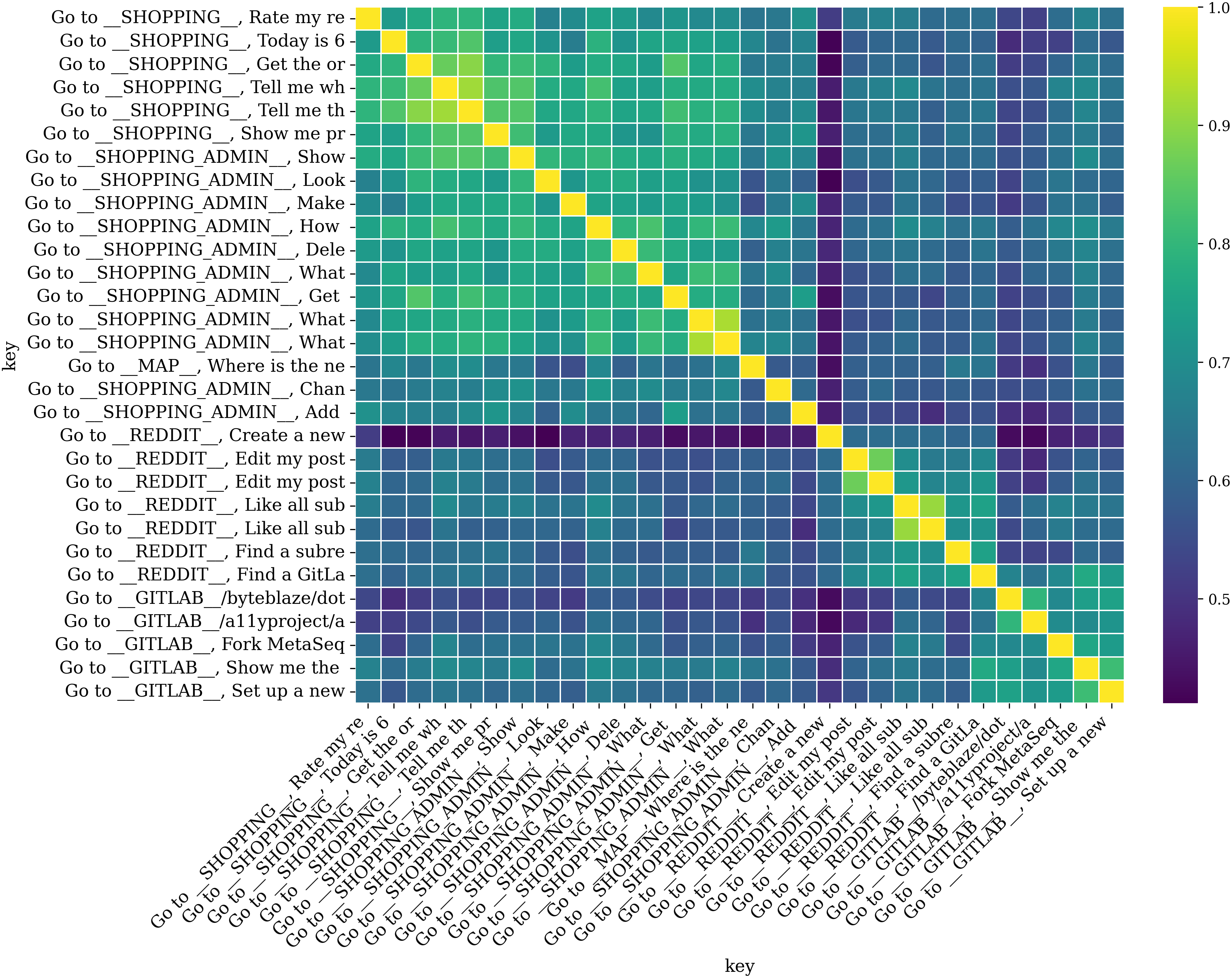}
    \caption{Cosine similarity heatmap for \texttt{llama8b-nnetnav-wa}, showing weaker separation between closely related categories.}
    \label{fig:nnetnav-heatmap}
\end{figure*}

\begin{figure*}
    \centering
    \includegraphics[width=0.75\textwidth]{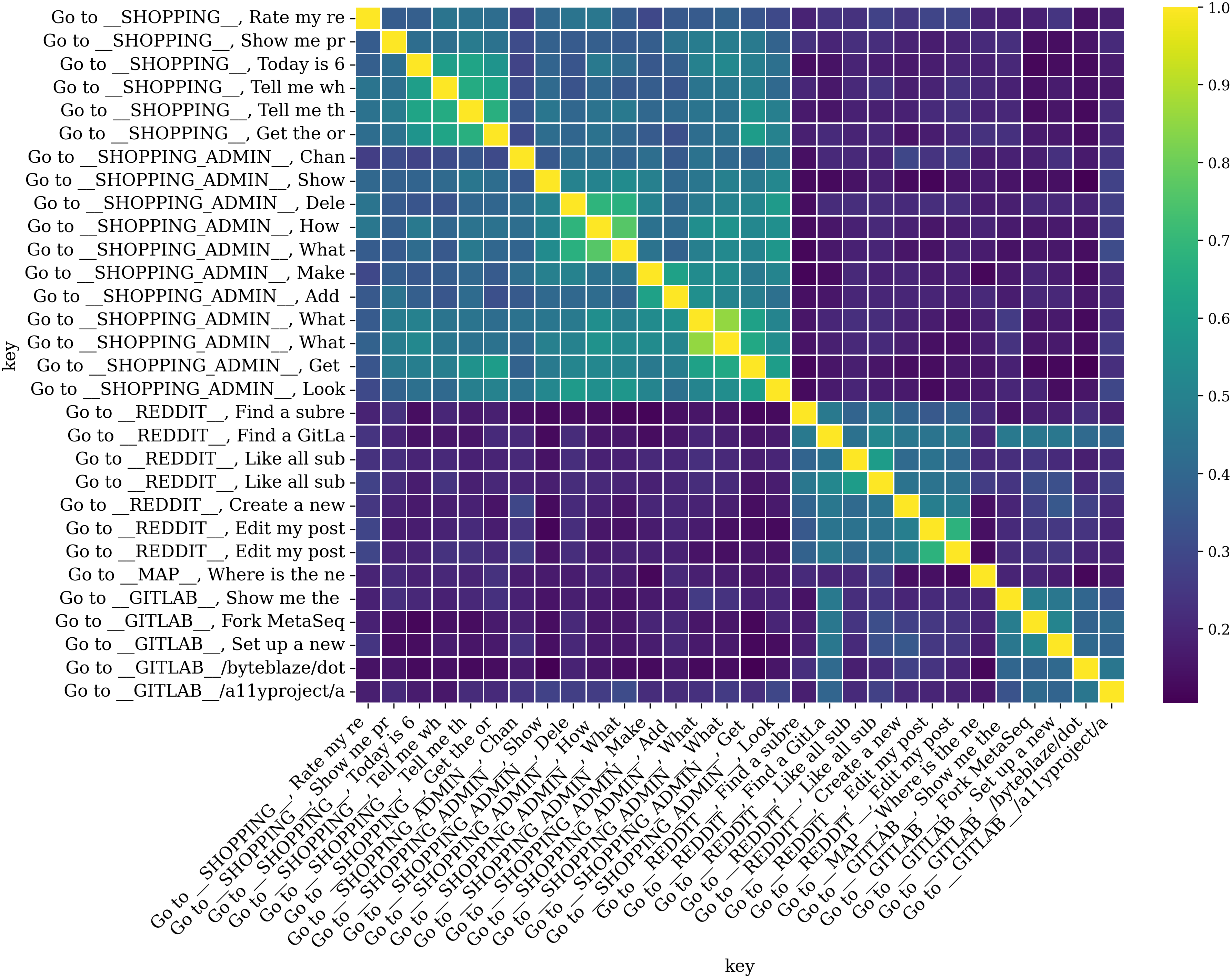}
    \caption{Cosine similarity heatmap for \texttt{text-embedding-3-large} across WebArena task categories.}
    \label{fig:openai-large-heatmap}
\end{figure*}

\begin{figure*}
    \centering
    \includegraphics[width=0.75\textwidth]{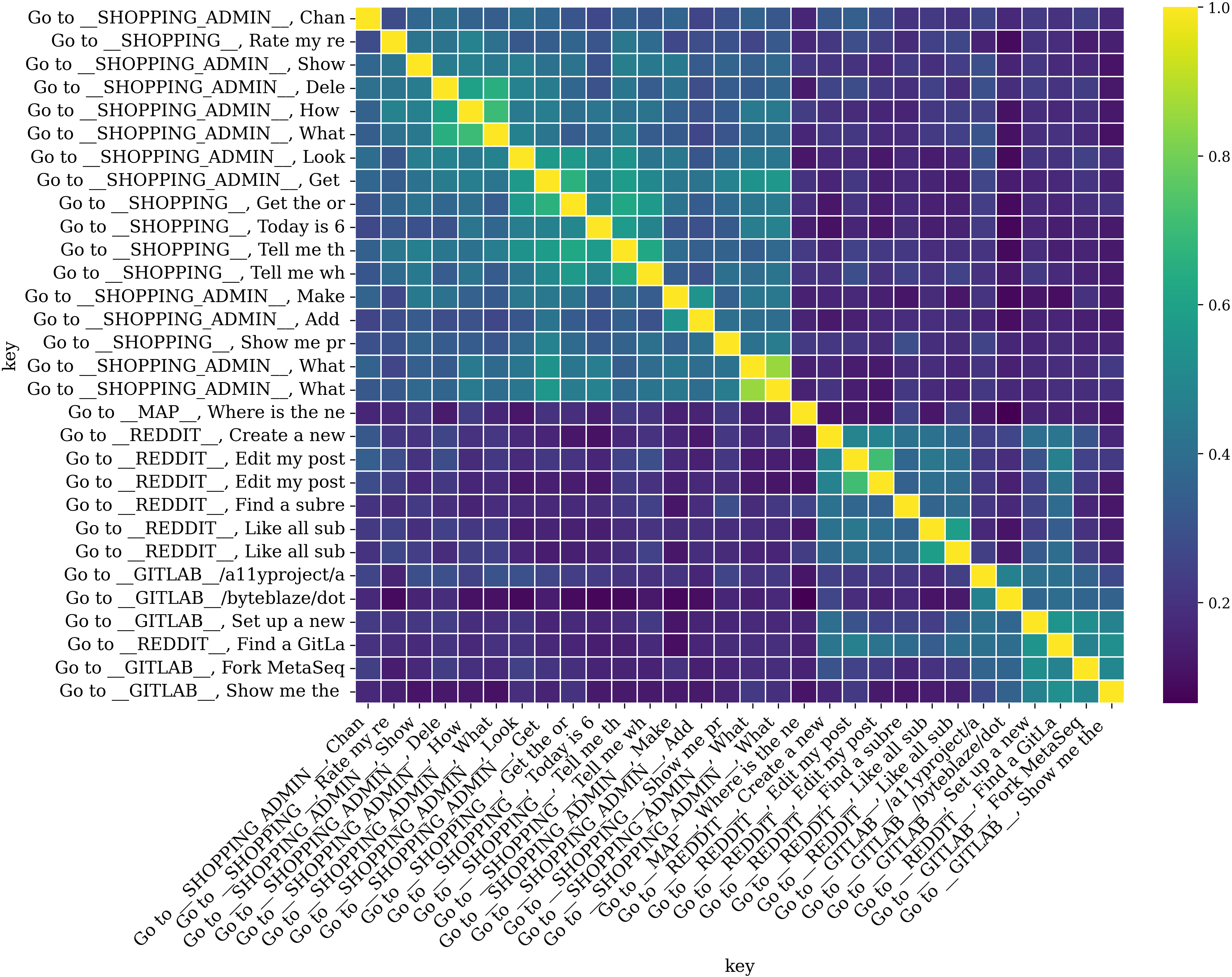}
    \caption{Cosine similarity heatmap for \texttt{text-embedding-3-small} across WebArena task categories.}
    \label{fig:openai-small-heatmap}
\end{figure*}

\begin{figure*}
    \centering
    \includegraphics[width=0.75\textwidth]{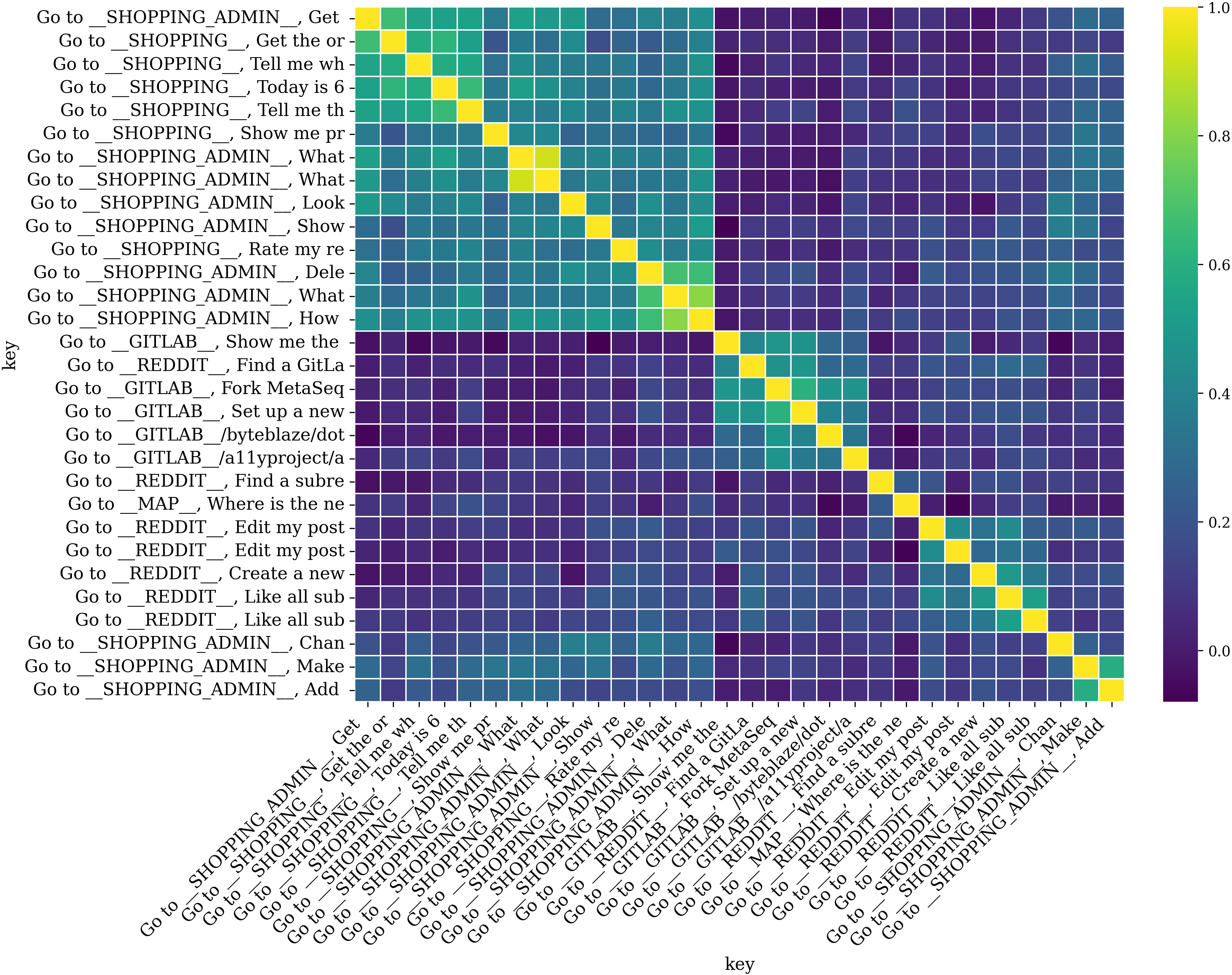}
    \caption{Cosine similarity heatmap for \texttt{all-MiniLM-L6-v2} across WebArena task categories.}
    \label{fig:all-MiniLM-L6-v2-heatmap}
\end{figure*}

Figures~\ref{fig:qwen-heatmap}–\ref{fig:all-MiniLM-L6-v2-heatmap} show cosine similarity matrices between task embeddings. From these visualizations, we observe that \texttt{llama8b-nnetnav-wa} (Figure~\ref{fig:nnetnav-heatmap}) struggles to separate \texttt{shopping} from \texttt{shopping\_admin} compared to larger models like \texttt{gte-Qwen2-7B-instruct} and \texttt{text-embedding-3-large}.

We select \texttt{gte-Qwen2-7B-instruct} as our embedding model for experiments in Section~\ref{sec:h2}. While it shows some difficulty distinguishing \texttt{shopping} and \texttt{shopping\_admin}, Figure~\ref{fig:site_spider_h2} shows that it can still retrieve helpful reflections. Notably, it contributes to an improvement of +12 points on both task categories.




\section{Data and Code Usage}
This work builds upon two key open-source codebases: WebArena~\cite{zhou2023webarena} and AgentOccam~\cite{yang2024agentoccam}, both of which are available under the Apache 2.0 License\footnote{\url{https://www.apache.org/licenses/LICENSE-2.0}}. In accordance with this license, we utilized WebArena as our evaluation environment and employed AgentOccam as our baseline agent. 

\end{document}